\documentclass[letterpaper]{article} % DO NOT CHANGE THIS
\usepackage{aaai25}
\usepackage{times}  % DO NOT CHANGE THIS
\usepackage{helvet}  % DO NOT CHANGE THIS
\usepackage{courier}  % DO NOT CHANGE THIS
\usepackage[hyphens]{url}  % DO NOT CHANGE THIS
\usepackage{graphicx} % DO NOT CHANGE THIS
\usepackage{natbib}  % DO NOT CHANGE THIS AND DO NOT ADD ANY OPTIONS TO IT
\usepackage{caption} % DO NOT CHANGE THIS AND DO NOT ADD ANY OPTIONS TO IT
\usepackage{algorithm}
\usepackage{algorithmic}
\usepackage{xcolor}         % colors
\usepackage{amsmath}
\usepackage{graphicx}
\usepackage{subfigure}
\usepackage{booktabs}       % professional-quality tables
\usepackage{amsfonts}
\usepackage{placeins}
\usepackage{newfloat}
\usepackage{listings}
\usepackage{fancyhdr}
\DeclareCaptionStyle{ruled}{labelfont=normalfont,labelsep=colon,strut=off} % DO NOT CHANGE THIS
\floatstyle{ruled}
\newfloat{listing}{tb}{lst}{}
\floatname{listing}{Listing}
\title{Learning Monocular Depth from Events via Egomotion Compensation}
\author{
    Haitao Meng, Chonghao Zhong, Sheng Tang, Lian JunJia, Wenwei Lin, Zhenshan Bing, Yi Chang, Gang Chen, Alois Knoll
}
\affiliations{
}

\nocopyright

\fancypagestyle{firstpage}{ 
	\fancyhf{} 
	\fancyfoot[L]{\scriptsize This work has been submitted to the IEEE for possible publication. Copyright may be transferred without notice, after which this version may no longer be accessible.}

}

\begin{document}
\thispagestyle{firstpage} 
\maketitle

\begin{figure*}[htbp]
	\centering	
	\includegraphics[width=1\textwidth, height=0.22\textheight]{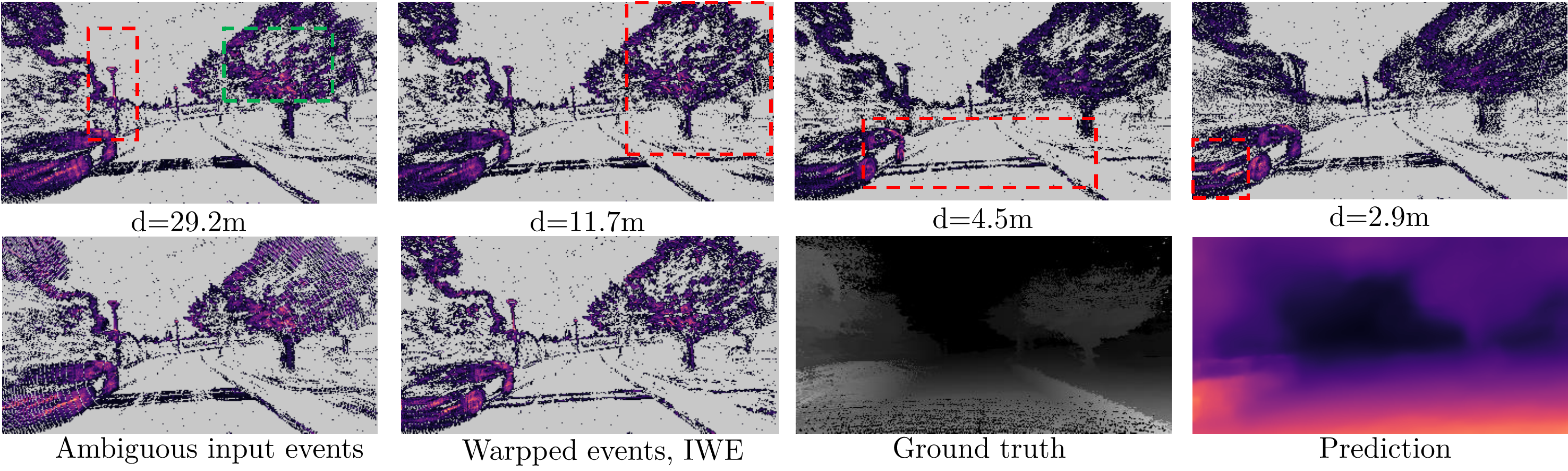}
	\caption{Image of wrapped event examples with different motion models under different depth hypotheses $d$, as well as the comparison of untouched ambiguous event image, warpped event image by using our depth prediction, the ground truth and our depth prediction. The \textcolor{red}{red} dashed boxes indicate the portion of IWE that are \textit{focused} (with the correct depth hypothesis). The \textcolor{green}{green} dashed box highlights the erroneously focused events caused by the repetitive texture of the tree. Better view in the color mode.}
	\label{fig:iwes}
%	\vskip -1cm
	\vspace{-0.5cm}
\end{figure*}
%\FloatBarrier

\begin{abstract}
Event cameras are neuromorphically inspired sensors that sparsely and asynchronously report brightness changes. Their unique characteristics of high temporal resolution, high dynamic range, and low power consumption make them well-suited for addressing challenges in monocular depth estimation (e.g., high-speed or low-lighting conditions). 
However, current existing methods primarily treat event streams as black-box learning systems without incorporating prior physical principles, thus becoming over-parameterized and failing to fully exploit the rich temporal information inherent in event camera data. 
To address this limitation, we incorporate physical motion principles to propose an interpretable monocular depth estimation framework, where the likelihood of various depth hypotheses is explicitly determined by the effect of motion compensation. To achieve this, we propose a Focus Cost Discrimination (FCD) module that measures the clarity of edges as an essential indicator of focus level and integrates spatial surroundings to facilitate cost estimation. Furthermore, we analyze the noise patterns within our framework and improve it with the newly introduced Inter-Hypotheses Cost Aggregation (IHCA) module, where the cost volume is refined through cost trend prediction and multi-scale cost consistency constraints. Extensive experiments on real-world and synthetic datasets demonstrate that our proposed framework outperforms cutting-edge methods by up to 10\% in terms of the absolute relative error metric, revealing superior performance in predicting accuracy.
\end{abstract}

\section{Introduction}
Event cameras are bio-inspired visual sensors that asynchronously respond to environmental changes. This characteristic of event cameras allows them to detect and record light intensity independently. 
Given an ideal event camera model, whenever the logarithmic brightness of a pixel $\{u_i, v_i\}$ alters exceeds the predefined threshold $\pm$C, event $e_i = (u_i, v_i, t_i, s_i)$ will be asynchronously generated. Compared to conventional cameras, event cameras have higher temporal resolution, higher dynamic range, and lower power consumption, which offer numerous opportunities and great importance for robust monocular depth estimation in autonomous robotics. 
However, the distinctive data acquisition mechanism of the event camera also poses a significant challenge in developing applicable algorithms. Due to its responsiveness to brightness changes, especially at edges, event cameras are highly sensitive to the movement of both objects and the camera platform itself. This sensitivity can cause the generation of event coordinates to shift for the same object over a short time interval. 
Consequently, the resulting image may manifest an ambiguous effect, as exemplified by the bottom left image in \mbox{Fig. \ref{fig:iwes}}.

Numerous efforts have been directed towards leveraging event data for monocular depth estimation tasks \cite{zhang2022spike, wang2021dual, gehrig2021combining, saxena2005learning}. However, these approaches often overlook the rich information of event data in the temporal domain. They treat the inherent ambiguity of event data as a limitation rather than a potential source of information. Consequently, these methods often achieve depth estimation at the cost of reduced accuracy or interpretability of the system.
For instance, several works in \cite{zhang2022spike, tian} introduce the powerful Transformer architecture to model event interaction within the temporal domain. Although they achieve solid prediction accuracy, the complex network is presented as a black box, making it difficult to understand and refine the underlying mechanisms. 
Alternatively, other researchers in \cite{zhu2019unsupervised,shi2023improved} attempt to pre-compensate for the motion of event cameras. They use predicted event-based optical flow maps to restore sharp event images and perform depth estimation. However, their system is susceptible to fluctuations due to the variability introduced by the additional motion prediction component. As a result, the prediction accuracy is unsatisfactory.
\textcolor{black}{
	To date, there has not been an extensive study on fully leveraging the unique temporal clues of event data for monocular depth estimation, leaving the following fundamental open question: how to exploit the potential of ambiguous event data and maximize its performance?}

In this paper, we aim to address this knowledge gap by distinctly learning and utilizing the inherent ambiguous effect of event data. With the incorporation of fundamental physical motion principles, we explicitly model the relationship between the motion of the event camera and the depth, achieving an interpretable and high-accuracy framework for monocular depth estimation.
Specifically, our proposed framework first exemplifies the event data into different measurable ambiguous image representations under different depth hypotheses. Then, we exploit the proposed Focus Cost Discrimination (FCD) module to extract the edge significance as the fundamental indicator and aggregate information from spatial surroundings to quantify the focus effect and facilitate the matching cost volume.
Finally, we introduce our Inter-Hypotheses Cost Aggregation (IHCA) module to address the noise patterns within the algorithm. This module analyzes cost trends across different depth hypotheses and imposes consistency constraints to multi-scale for the cost calculation, ultimately achieving reliable depth estimation results.

Our contributions are the following:
\begin{itemize}
	
	\item We propose a novel motion compensation-based monocular depth estimation framework for event cameras in which the depth is directly predicted with a metric scale.
	
	\item We propose an effective focus effect evaluation approach that measures different focus effect images to facilitate the cost volume. Additionally, we further analyze the noise patterns within our algorithm and develop a corresponding reduction module.
	
	\item  Evaluation of our method on public dataset and custom synthetic dataset both achieve leading accuracy when compared to state of the arts.
\end{itemize}

\section{Prior Work on Event-based Monocular Depth Estimation}

Monocular depth estimation, as a cornerstone task in computer vision, has been explored for decades\cite{saxena2005learning, zhang2018deep,  jiang2019hierarchical}. Despite the significant advancements facilitated by the advent of deep learning techniques, challenges persist in extreme scenarios, such as high-speed motion or low-light conditions \cite{zhu2019unsupervised, s23041998}. 
Recent studies have explored the potential of event cameras, attempting to leverage their unique characteristics in addressing these limitations in extreme scenes \cite{shi2023improved, learningdense}. However, the sparse and asynchronous nature of event cameras also poses new challenges for developing practical algorithms.

Although there has been a growing interest in monocular event cameras \cite{chaney2019learning, cm, haessig2019spiking, kim2016real, lowlatency}, these approaches are still limited by the achievement of only semi-dense or sparse depth estimation.
To produce dense depth prediction, \textit{Hidalgo-Carrio et al.}  \cite{learningdense} propose the first dense event-based depth prediction framework. They exploit ConvLSTM modules as the primary component, adapting a variant of the U-Net network for the prediction task. However, limitations arise due to the loss of temporal relationships in the event data during pre-processing, as highlighted by \textit{Shi et al.} \cite{shi2023improved}. As a result, the accuracy of this approach is significantly hampered.
%due to the loss of temporal relationship of event data in pre-processing \cite{shi2023improved}, its accuracy is significantly limited.
To enhance accuracy performance, \textit{Zhuang et al.}  \cite{zhang2022spike} introduce a powerful Transformer \cite{NIPS2017_3f5ee243} for the event-based depth estimation task.
Building upon this work, \textit{Liu et al.}  \cite{tian} further propose an enhanced architecture that embeds the stream data into the image-like format. 
However, while both approaches achieve promising results, they treat the event data similarly to traditional images, neglecting most of rich temporal information. Consequently, they failed to fully exploit the potential of the event data in the temporal domain.

Nevertheless, there are researchers that actively strive to leverage the distinctive properties of event data. \textit{Zhu et al.} \cite{zhu2019unsupervised} leverage a jointly trained neural network to predict the motion of the events for deblurring the image, aiming to achieve event images with sharp edges and rich texture. Similarly, \textit{Shi et al.} \cite{shi2023improved} propose directly inputting the optical flow map into the network and exploiting multiple flow compensation and correlation to generate the depth maps. 
\textit{Wang et al.} \cite{wang2021dual} claim to utilize the high dynamic range property of the event data to reconstruct the gray image. By employing sophisticated monocular estimators, they achieve promising performance in low-lighting conditions.
In contrast, \textit{Gehrig et al.} \cite{gehrig2021combining} addresses the same problem with the help of inputting gray images. They propose an asynchronous recurrent network module to extract the best from event data and gray images to integrate the final results.
The most related work to ours are \cite{chiavazza2023low} and \cite{zhu2018realtime}. In \cite{chiavazza2023low},  the authors define the camera movement as translational without any rotation, such that they can achieve sparse depth prediction based on dynamic motion equations.
In contrast, our work does not rely on any specific motion pattern. The proposed framework is capable of producing reliable dense depth maps in complex motion scenes.
Regarding the work in \cite{zhu2018realtime}, although we share the same depth-assumed motion compensation method, it is important to distinguish that \cite{zhu2018realtime} is a stereo-based method. It heavily relies on the consistency between stereo images to regulate its cost volume and produce sparse or semi-dense depth estimation. 
As a comparison, our work focuses on dense monocular depth estimation.  The underlying design principles for cost formulation and aggregation differ significantly.

\begin{figure*}[]
	\centering	
	\includegraphics[width=1\textwidth, height=0.5\textwidth]{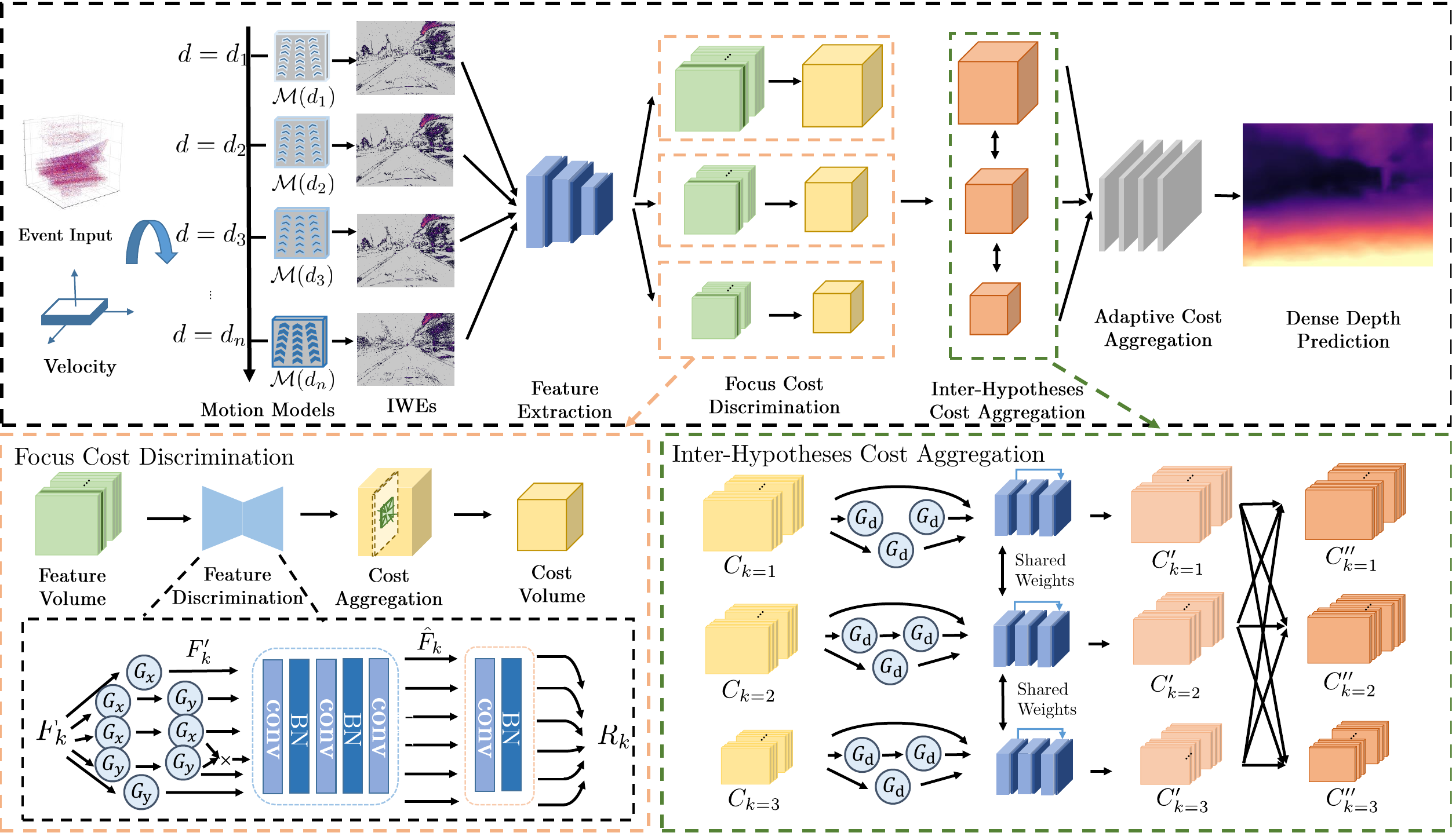}
	\caption{The overview of our proposed monocular event depth estimation framework. }
	\label{fig:overview}
		\vspace{-0.5cm}
\end{figure*}

\section{Method}

\subsection{Overview}

While prior research has explored various methods to compensate for camera dynamics and improve depth estimation accuracy, the use of over-parameterized sub-networks for camera motion prediction and depth estimation remains a concern due to limitations in interpretability and domain transferability \cite{tian, zhang2022spike, shi2023improved}.
To address this challenge,
we integrate a physical model into the monocular depth estimation task to simplify the complexities of the problem. 
Leveraging the classical dynamic motion field equation \cite{zhu2018realtime, zhu2019unsupervised,  chiavazza2023low}, we explicitly model the intricate relationship between the movement of the event camera on the 2D image (known as optical flow) and the depth of the individual pixel.
Subsequently, we utilize the Image of Warped Event (IWE), compensated by various depth-assumed optical flow models with different $focus$ level, to infer the final depth.  
To elaborate it in detail, we illustrate the overview of our proposed framework in \mbox{Fig. \ref{fig:overview}}. 

Beginning with the given events stream and measured velocity (e.g., from the Inertial Measurement Unit), we derive diverse optical flow models under different depth hypotheses to exemplify the measurable imagery agents. These agents are then directly concatenated along the channel dimension to facilitate the input of the network model. 
In the feature extraction module, we employ classical ResNet \cite{he2016deep, concentrate} and feature pyramid network \cite{psmnet, concentrate} to extract the features and hierarchically encode them at varying scales. 
To initiate the matching cost from the features across different depth hypotheses,  we propose a Focus Cost Discrimination module (FCD), in which different $focus$ levels in the imagery results will be quantified to scores, with the aim of indicating how well the depth-assumed motion model compensates for the movement of the camera and mitigates the $unfocused$ effect.
Similar to other monocular depth approaches \cite{gehrig2021combining, zhang2022spike}, we here explicitly formulate the cost volume for every candidate depth.  However, given that the depth hypotheses in our pipeline are in metric scale (from \mbox{Eq. \ref{dynamic_equation}}), rather than a relative scale to the image, we can thereby circumvent the scale ambiguity problem.
Subsequently, we further introduce an effective cost aggregation module. This module analyzes the trend of the cost across different depth hypotheses and leverages multi-scale information to promote the cost volume. 
Finally, we leverage the adaptive aggregation module \cite{aanet} and refinement module \cite{stereodrnet} to regulate and refine the final depth prediction.

\subsection{Egomotion Estimation Model}
\label{dynamic_motion_model}

As we stated earlier, the extent of the focus effect in the resulting IWE depends on the camera velocity and the distance to the image plane. This relationship can be modeled using the well-known dynamic field equation \cite{zhu2018realtime, zhu2019unsupervised} to provide insights into the behavior of the unfocused events over time.
Given a moving event camera with ideally measured translational velocity $ T=\{t_x, t_y, t_z\}$, rotational velocity $\omega = \{\omega_x, \omega_y, \omega_z\}$, and its intrinsic parameters, the velocity model $\mathcal{V}(d)$ for pixel $p_i =(u_i, v_i)$ can be defined under the depth hypothesis $d$:
\begin{equation}
	\label{dynamic_equation}
	\begin{aligned}
		\hspace{-0.2cm} 
		\mathcal{V}(d) &= 
		\frac{1}{d}
		\left[
		\begin{aligned}
			-f & \quad 0 \quad u_i^{'} \\
			0 & \quad -f \quad v_i^{'}
		\end{aligned}
		\right]
		T \\
		&\quad \hspace{0.7cm}
		+ \frac{1}{f}
		\left[
		\begin{aligned}
			u_i^{'}v_i^{'}  &\quad -(f^2 + u_i^{'2}) \quad v_i^{'} \\
			f^2+v_i^{'2} &\quad -u_i^{'}v_i^{'} \quad -fu_i^{'}
		\end{aligned}
		\right]
		\omega
	\end{aligned}
\end{equation}

\begin{equation}
	u_i^{'} = u_i - c_u , \qquad v_i^{'} = v_i - c_v \nonumber
\end{equation}

where $\Delta u_i(d)$ and $\Delta v_i(d)$ represent the x-axis and y-axis velocity component, respectively.  $f$ indicates the focal length and $c_u$ and $c_v$ are the optical centers of the camera.  With this dynamic model, we can formulate the motion model $\mathcal{M}(d)$ for warping the events set $E$ to the reference time $t_{ref}$ with the depth hypothesis $d$:
\begin{equation}
	\mathcal{M}(d) = \left(
	\begin{aligned}
		u_i^{\prime} \\
		v_i^{\prime}
	\end{aligned}
	\right)
	= \left(
	\begin{aligned}
		u_i \\
		v_i
	\end{aligned}
	\right)
	+ 
	\mathcal{V}(d)
	(t-t_{ref})
\end{equation}
where $u_i^{\prime}$ and $v_i^{\prime}$ are the warped coordinate of point $p_i$.
By selecting different depth hypotheses $d$, we can derive different dynamic models $\mathcal{M}$ for the events. However, only when the hypothesis is consistent with reality can allow the warped events to focus on the right positions. Incorrect depth hypothesis may further aggravate the unfocused effect and disturb the imagery result. To better illustrate this pattern, we present \mbox{Fig \ref{fig:iwes}} as an example to show the imagery effect under various motion models.

\subsection{Event Stacking}
\label{event_stacking}

For the stream event set $E$, we convert it into a 2D sparse image to avoid recurrent state encoding and propagation. 
Ideally, we expect the produced image to encompass the explicit feature of unfocused effect, with the aim of knowing how well the candidate ego-motion model fits the movement trajectory. Therefore, we propose representing the event data $E$ as the IWE $I(\mathcal{M}(d))$ by quantifying the occurrences of events at each pixel under the depth-assumed motion model $\mathcal{M}(d)$. 
%The conversion can be parameterize as:
\begin{equation}
	I(p;\mathcal{M}(d)|) = \sum\nolimits_{i=1}^{n}\mathbf{1}_{(p_i=p)}
\end{equation}
Different from \cite{zhu2018realtime}, our accumulation omits the polarity property of events. This divergence arises from our design principle of amplifying the \textit{focus} gap between the IWEs by correct and incorrect motion models. The overlap of positive and negative polarity data effectively doubles the stacking of the events at pixels, which helps us to discriminate the representation $I$ from different candidate motion models $\mathcal{M}$. 

\subsection{Focus Cost Discrimination}
\label{focus_cost_discrimination}

Following the accumulation of the events data $E$ into image representations $I(\mathcal{M}(d))$ and their subsequent feature encoding, our objective shifts towards evaluating the focus effect and quantitatively measuring it as a score to integrate the matching cost volume.

While sophisticated works leverage Contrast Maximization (CM) methods as the theoretical foundation, employing deformed energy objective functions to infer the clarity of the images \cite{secrets, zhu2019unsupervised, Stoffregen19cvpr}, these works, however, are either noise-sensitive or prone to overfitting to specific data \cite{cm, secrets}, and thereby are unable to support conducting precise focus assessment.

In our work, we hold the assumption that focused images tend to exhibit sharper edges as the underlying principle for developing our module. For the  $k^{th}$ input scale of feature volume $F_k(\mathcal{M}(d))$, we derive the first order gradient $ \{ G_x, G_y \}$, the second order gradient $ \{ G_{xx}, G_{yy}, G_{xy}\}$, and their combination $\{ G_{xx}*G_{yy}\}$ as the indicator for the significance of the edges, obtaining the transformed features volume $F^\prime_k(\mathcal{M}(d))$. Then, we craft a feature aggregation block $f$ to adaptively gather the gradient information, as shown in the subfigure (a) of \mbox{Fig. \ref{fig:overview}}. To prevent the proposed block from being over-parameterized, the convolution layers are configured as channel-separated, with the aim of diminishing the disturb from the gradient maps of other depth candidate hypotheses.
\begin{align}
	\hat{F}^X_k(\mathcal{M}(d)) = f ( F^\prime_k(\mathcal{M}(d); G_{X})) \qquad \qquad \qquad \\ \nonumber
	\text{where} \quad X= \{x, y, xx, yy, xy, xx*yy\}
\end{align}
The final gradient representation $R$ is defined as the combination of all gradient maps corresponding to the same depth hypothesis $d$ with learnable weights $w$.
\begin{align}
	R(\mathcal{M}(d)) = \sum^{X} w_X \hat{F}_k^X(\mathcal{M}(d))
\end{align}
After that, we have finished the significant feature extraction and denoise for the input IWEs. To obtain the likelihood cost of each depth candidate $d$ and form the overall matching cost volume, we simply aggregate the pixels in gradient representation map $R(\mathcal{M}(d))$ by leveraging the support of surrounding region $S$ with the size of $r \times r$.
\begin{align}
	C(\mathcal{M}(d)) = \sqrt{\sum_{p\in S_{r\times r}} (R(p;\mathcal{M}(d))^{2}} 
\end{align}

\begin{figure*}[]
	\centering
	\subfigure[E2D \cite{learningdense}]{
		\begin{minipage}[t]{0.32\linewidth}
			\centering
			\includegraphics[width=1\linewidth,height=1.55\columnwidth]{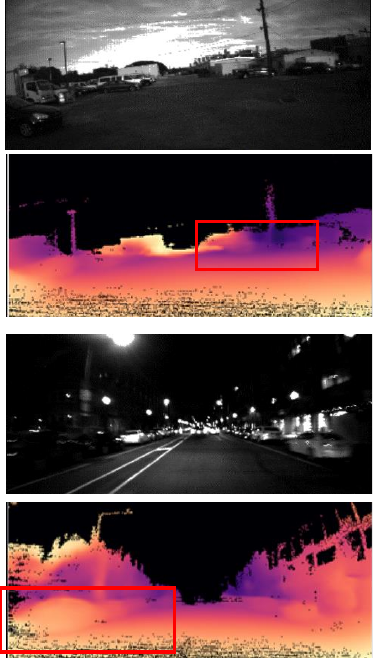}
			\label{fig:ad}
		\end{minipage}%	
	}
	\subfigure[EReFormer \cite{tian}]{
		\begin{minipage}[t]{0.32\linewidth}
			\centering
			\includegraphics[width=1\linewidth,height=1.55\columnwidth]{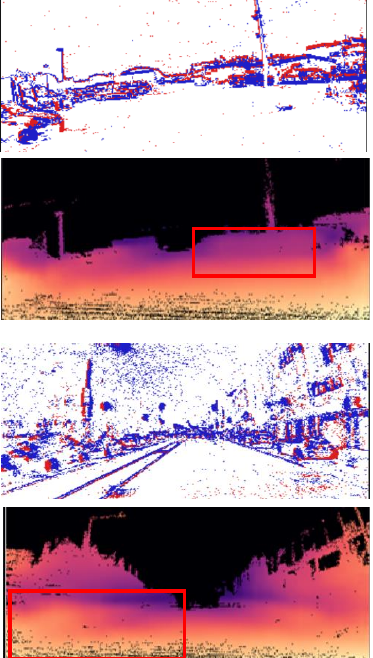}
			\label{fig:ad}
		\end{minipage}%	
	}
	\subfigure[Ours]{
		\begin{minipage}[t]{0.32\linewidth}
			\centering
			\includegraphics[width=1\linewidth,height=1.55\columnwidth]{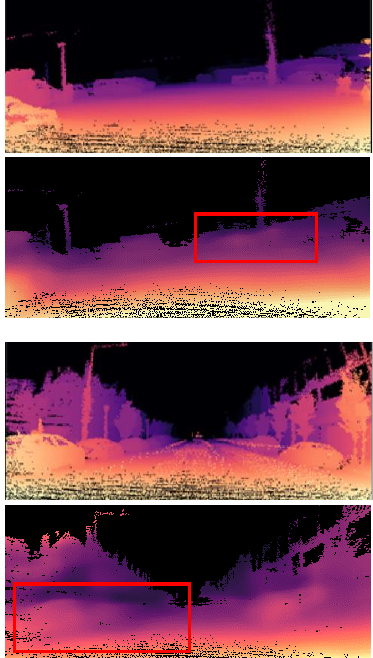}
			\label{fig:ad}
		\end{minipage}%	
	}
	\centering
	\vskip -0.35cm
	\caption{ The representative examples of our model in comparison with other state of the arts. The first and the third rows illustrate the APS image, the event data (converted as colored event image), and the ground truth depth map. The second and the fourth rows indicate the depth estimations of APS+IEBins \cite{shao2023IEBins}, E2D\cite{learningdense}, EReFormer \cite{tian}, and ours, respectively.}
	\vskip -0.2cm
	\label{fig:qua_mvsec}
\end{figure*}

\subsection{Inter-hypotheses Cost Aggregation}
\label{trend-aware}
In the above section, we have introduced an effective module for interpreting the unfocused imagery representation into comparable continuous cost. However, given that the focus gap derived from different motion models can be difficult to discern, accompanied by the potential mismatch cost calculation, we can only view the produced cost volume as a coarse cost volume. To further promote it, we take the trend of focus effect from different motion models into consideration and integrate multi-scale cost volume to mitigate the ambiguity of the cost.  

According to the \mbox{Eq. \ref{dynamic_equation}}, the velocity $\mathcal{V}(d)$ is monotonic negative correlation with the depth hypothesis $d$. Given time interval $t-t_{ref}$ for events $E$ to the reference time is fixed, the optical flow model $\mathcal{M}(d))$ thereby can be seen as a monotonic negative correlation with depth candidate $d$.  That is, a smaller $d$ will lead to a greater change in event coordinates. However, only when the depth candidate aligns precisely with the ground truth, can the offset in coordinates result in a focused image. Neither excessively large nor small offsets can aggregate event trajectories. Therefore, the focus effect in our context exhibits a clearly unimodal property. See \mbox{Appendix A} for additional unimodal samples. To utilize this characteristic, we learn the trend of the costs for pixels among the depth hypotheses to refine the cost generation.

For the generated cost volume from Focus Cost Discrimination module $C(\mathcal{M}(d)) \in \mathbb{R}^{B\times D \times H \times W}$, 
we explicitly extract trend features using hand-crafted first-order and second-order gradients along the $D$ dimension \{$G_{d}, G_{dd}$\}, which inherently capture the fluctuating nature of the cost.  Subsequently, we associate the spatial information along $H$ and $W$ dimensions with these trend features to jointly analyze the cost trend. 
Subfigure (b) in \mbox{Fig. \ref{fig:overview}} illustrates the architecture of the proposed module, where stacked residual 3D convolution blocks are employed to facilitate weight learning and to regenerate the cost volume as $C^\prime$.

Meanwhile, we observe that event stacking can sometimes lead to a misleading local optimal focus effect, even when the depth hypothesis deviates from the ground truth. We found this phenomenon arises due to the coincidental overlap of events caused by repetitive textures, as illustrated by the green box in \mbox{Fig. \ref{fig:iwes}}
To diminish this noise, we leverage lateral knowledge from the multi-scale cost volume, enforcing cost consistency across various scales. The definition can be summarized as:
\begin{align}
	C^{\prime\prime}_k(\mathcal{M}(d)) = \sum\nolimits^{n} f_k (C^\prime(\mathcal{M}(d))) 
\end{align}
where $f$ is the neural network layer and $k$ represents the $k^{th}$ scale.

\begin{table*}[]
	
	\centering
	\caption{Quantitative evaluation on MVSEC dataset. $\downarrow$ indicates the the lower value is preferred, while $\uparrow$ signifies the opposite. The best is in $\textbf{bold}$ and the running up is \underline{underlined}. }
	\resizebox{1\textwidth}{!}{
		\begin{tabular}{cccccccccccc}
			\toprule
			Dataset & Method & Abs.Rel $\downarrow$ & Sq.Rel$\downarrow$ & RMSE$\downarrow$ & RMSE log $\downarrow$ &
			$\delta$\textless{}1.25 $\uparrow$ & $\delta$\textless{}$1.25^2$ $\uparrow$ 
			&$\delta$\textless{}$1.25^3$ $\uparrow$ 
			&10m $\downarrow$
			&20m $\downarrow$
			&30m $\downarrow$
			\\  \midrule
			
			&  E2D \cite{learningdense} &  0.450 & 0.627  & 9.321 &  0.514   &  0.472    &    0.711       & 0.823 &2.70&3.46&3.84 \\
			&DTL \cite{wang2021dual}&0.390&---&---&0.436&0.510&0.757&0.865 &2.00&2.91&3.35\\
			Day&EF2DNet \cite{shi2023improved}  &0.319  &0.553&\underline{8.333} &0.389 &0.600&0.799 &0.897&1.50&2.39&2.91\\
			& E2D+ \cite{learningdense}&  0.346  & \underline{0.516} & 8.564 & 0.421  &  0.567 &  0.772 & 0.876&1.85&2.64&3.13\\
			&EReFormer \cite{tian}&\underline{0.271}&---&---&\underline{0.333}&\underline{0.664}&\underline{0.831}&\underline{0.923}&\underline{1.29}&\underline{2.14}&\underline{2.59}\\
			& Ours   & \textbf{0.223} & \textbf{0.247}  & \textbf{7.182}& \textbf{0.312}   &\textbf{0.708} & \textbf{0.865} &\textbf{0.939} &\textbf{0.88}& \textbf{1.68}&\textbf{2.12} \\
			\midrule
			
			& E2D\cite{learningdense}  & 0.770  &3.133& 10.548 & 0.638 & 0.327& 0.582  & 0.732 & 5.36& 5.32 &5.40 \\
			&DTL\cite{wang2021dual}&0.474&---&---&0.555& 0.429& 0.657&0.791&2.61& 3.11& 3.82\\
			Night&EF2DNet \cite{shi2023improved}&0.428&\underline{1.781}&\underline{8.869 }&0.467 &0.529&0.725 &0.849 &2.16&2.91&3.43 \\
			& E2D+\cite{learningdense}& 0.591&2.121&11.210&0.646 &0.408&0.615&0.754&3.38& 3.82&4.46 \\
			&EReFormer \cite{tian}&\textbf{0.317}&---&---&\underline{0.415} &\underline{0.547}&\underline{0.753}&\underline{0.881} &\textbf{1.52}&\textbf{2.28}&\textbf{2.98}\\
			&  Ours   & \underline{0.367} & \textbf{1.573 } & \textbf{8.141}& \textbf{0.412}   &\textbf{0.554} & \textbf{0.766} &\textbf{0.886} &\underline{1.76}&\underline{2.77}&\underline{3.23} \\
			\bottomrule
		\end{tabular}
	}
	\label{tab:mvsec1}
\end{table*}

\section{Experiment}

\subsection{Experimental Setup}
\textbf{Datasets.} We evaluate our approach on two datasets: the Multi-Vehicle Stereo Event Camera (MVSEC) dataset \cite{mvsec} and EventCitySim. MVSEC is a popular dataset comprising several sequences covering day and night-time driving scenarios. EventCitySim is a newly released synthetic dataset that we generated in the CARLA simulator \cite{carla} using the event camera plugin \cite{learningdense}. It includes 5 sequences captured under different lighting conditions, with a total of 7500 synchronized RGB images, depth maps, and continuously recorded event data, IMU measurements, and gyroscope data.  The detailed information and image examples of EventCitySim can be found in \mbox{Appendix B} .

\textbf{Evaluation Metrics.} The comparison in our experiments contains various metrics \cite{zhang2022spike, learningdense, tian}, including absolute relative error (Abs.Rel), square relative error (Sq.Rel), root mean square error (RMSE) and its logarithm version (RMSE log), accuracy ($\delta<1.25^n$, n=1,2,3), the average absolute depth error at 10m, 20m, and 30m cutoff distance and end point errors (EPE). The detailed formulations of these metrics can be found in  \mbox{Appendix C}.

\textbf{Implementation Details.} 
We built our model using the PyTorch framework. 
Following \cite{zhu2018realtime}, we leverage the translational and rotational velocities that are interpolated by using odometry poses to construct our motion model.
For the model training, we set the learning rate to 0.001 and apply a weight decay of 0.0001.
We use 80,000 events for the MVSEC dataset and 100000 events for the EventCitySim dataset to initial the input formulation, alone with a maximum time interval limitation of 0.2 seconds.
We train our model using a dual Titan RTX GPU server equipped with an Intel i7-9800X CPU. The model is trained for a total of 30 epochs with a batch size of 4.

\subsection{Results}
\textbf{Performance Comparison on MVSEC dataset.} We first conduct the result comparison with state-of-the-art depth estimators on the MVSEC dataset. For the sake of fairness, we follow the work in  \cite{zhu2018realtime, tian, learningdense} to train our model on the sequence \textit{outdoor\_day2} and report the results evaluated on the sequence \textit{outdoor\_day1} and \textit{outdooor\_night1}. 

The performance of our framework is evaluated using various metrics on both daytime and nighttime sequences. \mbox{Tab. \ref{tab:mvsec1}} shows the comparison results.  
Overall, Our framework achieves the highest accuracy across all metrics in daytime sequences and ranks among the top six performers in nighttime sequences when compared to all state-of-the-art methods. We attribute this superiority to the incorporation of the physical model, which provides anchoring depth hypotheses that mitigate the scale ambiguity commonly encountered by other approaches.

\begin{table*}[]
	
	\centering
	\caption{Quantitative evaluation on sequence \textit{town4} and \textit{town5\_night} of EventCitySim dataset. The best is in $\textbf{bold}$ and the running up is \underline{underlined}. 
		It is important to note that depths exceeding 80 meters are excluded from this comparison since events are rarely triggered beyond such distances. }
	\resizebox{1\textwidth}{!}{
		\begin{tabular}{cccccccccccc}
			\toprule
			Dataset & Method & Abs.Rel $\downarrow$ & Sq.Rel$\downarrow$ & RMSE$\downarrow$ & RMSE log $\downarrow$ &
			$\delta$\textless{}1.25 $\uparrow$ & $\delta$\textless{}$1.25^2$ $\uparrow$ 
			&$\delta$\textless{}$1.25^3$ $\uparrow$ 
			&10m $\downarrow$
			&20m $\downarrow$
			&30m $\downarrow$
			\\  \midrule

			&E2D \cite{learningdense} &\underline{0.402}&0.560&\underline{10.798}&0.557&0.230&0.604&0.788&\textbf{1.845}&\textbf{3.132}&\underline{4.156}\\
			town4&EReFormer \cite{tian}&0.480&\underline{0.540}&\textbf{10.786}&\underline{0.506}&\underline{0.363}&\underline{0.66}9&\underline{0.833}&\underline{2.014}&5.06&5.99\\
			 &Ours & \textbf{0.376}&\textbf{0.426}& 11.344& \textbf{0.453} &\textbf{0.528} &\textbf{0.714}&\textbf{0.836}&2.160&\underline{3.298}&\textbf{3.387} \\
			
			\midrule
			
			&E2D \cite{learningdense} &0.539&1.650&\textbf{13.615}&\underline{0.633}&\underline{0.361}&\underline{0.675}&\textbf{0.781}&{2.813}&\underline{3.361}&\underline{4.111}\\
			town5&EReFormer \cite{tian}&\underline{0.484}&\underline{0.542}&\underline{14.775}&0.716&0.335&0.610&0.752&\underline{2.49}&4.07&5.52 \\	
				&Ours&\textbf{0.319}&\textbf{0.252}& {15.345}&\textbf{ 0.533} &\textbf{0.503} &\textbf{0.678}&\textbf{0.781}&\textbf{1.317}&\textbf{1.708}&\textbf{2.580} \\
			\bottomrule
		\end{tabular}
	}
	\label{tab:eventcitysim}
\end{table*}

\textbf{Performance Comparison on EventCitySim dataset.} 
Additional dataset evaluation is performed on the proposed simulated dataset. For the total 5 sequences (\textit{town1, town2\_night, town3, town4, town5\_night}), we utilize the first three for training and the rest two for validation, enabling the comprehensive training and testing on both day and night scenes.
The quantitative results of E2D \cite{learningdense}, EreFormer \cite{tian} and ours are illustrated in \mbox{Tab. \ref{tab:eventcitysim}}.
Overall, our approach achieves leading performance on two validation sequences. 

Specifically, we observe that our approach outperforms others significantly in terms of cutoff errors, especially as the distance increases. The 30m cutoff distance error of our framework on town4 is 2.58, whereas the two competing methods achieve errors of 4.11 and 5.52, respectively. We believe this significant difference arises from the introduction of the physical motion equation. Our approach leverages this equation to estimate depth at a metric scale, rather than a relative scale to the image, which contributes to the improved performance.

\textbf{Qualitative Examples.} Additional qualitative results of monocular depth estimators are shown in \mbox{Fig. \ref{fig:qua_mvsec}}. We compare the output of our approach against state-of-the-art monocular methods on both day and night scenes. Notably, even on unseen nighttime scenes (our model trained only on daytime sequences), it delivers reliable and sharp depth predictions.

%\subsection{Ablation Study.}
\textbf{Effect of proposed modules.} To validate the effectiveness of the proposed Focus Cost Discrimination module (FCD) and Inter-Hypotheses Cost Aggregation  (IHCA) module, we conduct an ablation study on the MVSEC dataset (sequence \textit{outdoor\_day2} for training and sequence \textit{outdoor\_day1} for validation).
For assessing the performance of the FCD module, we primarily compared it with four different cost quantification functions: squared timestamp images objective (sti) \cite{zhu2019unsupervised}, sum of suppressed accumulations objective (sosa) \cite{Stoffregen19cvpr}, sum of exponentials objective (soe) \cite{Stoffregen19cvpr}, and variance objective (var) \cite{gallego2017accurate}. 

For the IHCA module, we compared the accuracy of models with and without the IHCA module to assess the performance gap and verify its effectiveness.

\mbox{Tab. \ref{tab:ablation_fcd}} presents a comprehensive comparison of metrics. From this table, we can observe that $N_{\text{FCD}}$+IHCA (the default configuration of our proposed network) achieves the highest accuracy across all evaluated metrics.
In contrast, disabling the IHCA module ($N_{\text{FCD}}$) results in a notable drop in accuracy. This can be seen as strong evidence of the effectiveness of the IHCA module.
For the evaluation of the effectiveness of FCD module, we compare the metrics with four different cost quantification functions ($N_{\text{sti}}, N_{\text{sosa}}, N_{\text{soe}}$, and $N_{\text{var}}$).
It is clear from \mbox{Tab. \ref{tab:ablation_fcd}} that although they can adequately reconstruct dense depth maps within our proposed framework, none of them achieve the same level of high-quality estimation as our introduced FCD module. This highlights the superior performance and effectiveness of the proposed FCD module.

\begin{table}[]
	\centering
	\caption{ Accuracy comparison of employing different modules. $N$ refers to our proposed network framework, with suffixes indicating the specific cost function utilization. Note that $N_{FCD}$+IHCA refers to our proposed network configuration.
	}
	\resizebox{0.48\textwidth}{!}{
		\begin{tabular}{cccccc}
			\toprule
			& Ab.Rel$\downarrow$  & RMSE $\downarrow$ & $\delta<1.25^2$ $\uparrow$ & 20m $\downarrow$ &  EPE$\downarrow$ \\ 
			\midrule
				$N$$_{\text{FCD}}$ &   \underline{0.242 } &  \underline{7.332} &  {0.834} & \underline{1.715}& \underline{3.54}\\
				$N$$_{\text{FCD}}$+IHCA &\textbf{0.223}&\textbf{7.182}&\textbf{0.865} &\textbf{ 1.68} & \textbf{ 3.42}  \\
			
			\midrule

			$N$$_{\text{sti}}$\cite{zhu2019unsupervised} &  0.252 & 7.404    &  \underline{ 0.839}  &1.784 & 3.60 \\

			$N$$_{\text{sosa}}$ \cite{Stoffregen19cvpr} &0.421&9.824&0.726& 3.51&6.01\\
			
			$N$$_{\text{soe}}$ \cite{Stoffregen19cvpr} &0.268&7.582&0.829&1.969 & 3.74\\
			
			$N$$_{\text{var}}$ \cite{gallego2017accurate} &0.283&7.534&0.834&2.071&3.76\\

			\bottomrule
	\end{tabular}
}
	\label{tab:ablation_fcd}
\end{table}

\textbf{Impact of Velocity Noise.}
%\textcolor{red}{
For real-world applications, ensuring an accurate and unbiased velocity measurement is often challenging. Therefore, it is important to investigate the accuracy performance of our framework under various noise levels. In this ablation study, we artificially add varying levels of noise to the linear and angular velocity information inputted to our network, and report quantitative accuracy metrics on the MVSEC dataset. Following \cite{zhu2018realtime}, We independently generate noise for the three linear velocity components and the three angular velocity components. The noise follows a zero-mean Gaussian distribution, with the variance specified as a given percentage of the norm of the linear and angular velocity vector.  
For the case where multiple velocity measurements exist within the event stacking interval, we calculate the norm for the noise distribution by averaging the norms of the velocity vectors across the entire time interval.
\mbox{Tab. \ref{tab:ablation_noise}} presents the accuracy under various noise levels. It is noteworthy that our model exhibits robust resistance to velocity noise. Even with 100\% velocity noise, it maintains high-quality and reliable depth prediction outputs. The performance at this level of noisy velocity measurement is comparable to E2D+ \cite{learningdense} and EF2DNet \cite{shi2023improved} (as it is shown in \mbox{Tab. \ref{tab:mvsec1}}). We attribute this capability to the adoption of multi-scale consistency regulation in the IHCA module, which optimizes the cost trend to minimize the impact of noisy velocity.

\begin{table}[]
	\centering
	\caption{Ablation study of input velocity noise. The model is trained on \textit{outdoor\_day2} sequence by the velocity information without noise, and evaluate with different noisy input on \textit{outdoor\_night1} sequence. }
		\vskip-0.3cm
	\resizebox{0.48\textwidth}{!}{
		\begin{tabular}{ccccccc}
			\toprule
			& Abs.Rel $\downarrow$ & Sq.Rel$\downarrow$ & RMSE $\downarrow$
			 & $\delta$\textless{}$1.25^2$ $\uparrow$
			&20m$\downarrow$ & EPE$\downarrow$

			\\  \midrule

			No Noise  & {0.223} & {0.247}  &{7.182}& {0.865} &1.68& 3.42 \\
			\midrule
	
			10\% &0.232& 0.249&7.454& 0.854 & 1.73&3.57\\
			20\% &0.248& 0.271&7.744& 0.836& 1.82 & 3.76\\
			50\% & 0.286& 0.314& 8.246&0.799& 2.06&4.14\\
			100\% &0.313&0.361& 8.614& 0.776& 2.199& 4.395\\
			
			\midrule
			
		\end{tabular}
	}
	\vskip-0.3cm
	\label{tab:ablation_noise}
\end{table}

\section{Conclusion and Future Work}
This work introduces an effective approach for event-based monocular dense depth estimation. 
By integrating velocity information of the event camera, we achieve leading accuracy when compared to state-of-the-art methods across multiple datasets.
In this work, we propose a learning-based focus effect discrimination module, and provide detailed analysis and solution for potential noise within  our framework.  
The comprehensive experiments demonstrate that our method is accurate and reliable in both day and night scenes. Furthermore, with rigorous ablation study, we thoroughly demonstrate and analyze the effectiveness of the proposed modules and the robustness for the input velocity of our system.

On the other hand, we acknowledge that this work remains ineffective for static camera setups, due to the specific requirements for the egomotion of the event camera.
Therefore, in future work, we expect further research on this limitation. A potential solution could involve planning a regular cyclic motion for the event camera. Specifically, the cyclic motion should generate linear velocity measurements, as depth corresponds only to the linear velocity component, as shown in Eq. \ref{dynamic_equation}.

\bibliography{Bibliography}

\end{document}